\documentclass[conference]{IEEEtran}
\IEEEoverridecommandlockouts
\usepackage{cite}
\usepackage{amsmath,amssymb,amsfonts}
\usepackage{algorithmic}
\usepackage{graphicx}
\usepackage{textcomp}
\usepackage{xcolor}
\usepackage{soul}
\usepackage{tikz}
\newcommand{\mdl}[1]{{\sffamily \slshape \footnotesize #1}}
\newcommand{\expv}{\mathrm{E}}
\def\BibTeX{{\rm B\kern-.05em{\sc i\kern-.025em b}\kern-.08em
    T\kern-.1667em\lower.7ex\hbox{E}\kern-.125emX}}

\newcommand\copyrighttext{%
  \footnotesize \textcopyright 2020 IEEE. Personal use of this material is permitted. Permission from IEEE must be obtained for all other uses, in any current or future media, including reprinting/republishing this material for advertising or promotional purposes, creating new collective works, for resale or redistribution to servers or lists, or reuse of any copyrighted component of this work in other works. DOI: 10.1109/IJCNN48605.2020.9207573.}
\newcommand\copyrightnotice{%
\begin{tikzpicture}[remember picture,overlay]
\node[anchor=south,yshift=10pt] at (current page.south) {\fbox{\parbox{\dimexpr\textwidth-\fboxsep-\fboxrule\relax}{\copyrighttext}}};
\end{tikzpicture}%
}

\begin{document}

\title{Forecasting Photovoltaic Power Production using a Deep Learning Sequence to Sequence Model with Attention\\
\thanks{Support provided by the Future Energy Systems research initiative under the Canada First Research Excellence Program (CFREF) and grant RGPIN-2017-05866 from the Natural Sciences and Engineering Research Council (NSERC) of Canada are gratefully acknowledged. Titan Xp GPUs that enabled this research were provided by NVIDIA. }
}

\author{\IEEEauthorblockN{Elizaveta Kharlova*}
\IEEEauthorblockA{\textit{Electrical and Computer Engineering} \\
\textit{University of Alberta}\\
Edmonton, Canada \\
kharlova@ualberta.ca}
\and
\IEEEauthorblockN{Daniel May*}
\IEEEauthorblockA{\textit{Electrical and Computer Engineering} \\
\textit{University of Alberta}\\
Edmonton, Canada \\
dcmay@ualberta.ca}
{\footnotesize \textsuperscript{*} Authors with equal contribution, in alphabetical order}
\and
\IEEEauthorblockN{Petr Musilek}
\IEEEauthorblockA{\textit{Electrical and Computer Engineering} \\
\textit{University of Alberta}\\
Edmonton, Canada \\
Petr.Musilek@ualberta.ca}
}

\maketitle

\copyrightnotice

\begin{abstract}
Rising penetration levels of (residential) photovoltaic (PV) power as distributed energy resource pose a number of challenges to the electricity infrastructure. High quality, general tools to provide accurate forecasts of power production are urgently needed. In this article, we propose a supervised deep learning model for end-to-end forecasting of PV power production. The proposed model is based on two seminal concepts that led to significant performance improvements of deep learning approaches in other sequence-related fields, but not yet in the area of time series prediction: the sequence to sequence architecture and attention mechanism as a context generator. 

The proposed model leverages numerical weather predictions and high-resolution historical measurements to forecast a binned probability distribution over the prognostic time intervals, rather than the expected values of the prognostic variable. This design offers significant performance improvements compared to common baseline approaches, such as fully connected neural networks and one-block long short-term memory architectures. Using normalized root mean square error based forecast skill score as a performance indicator, the proposed approach is compared to other models. The results show that the new design performs at or above the current state of the art of PV power forecasting. 
\end{abstract}


\section{Introduction}
The shift towards more distributed energy resources (DER) and their subsequent high penetration levels negatively influence the electric infrastructure, e.g. through the duck curve problem~\cite{duck2015}. This results in a demand for accurate DER forecasting tools, especially for resources with high transient speeds and significant intermittency, such as photovoltaic (PV) solar energy.
At the same time, continuing progress towards the implementation of the smart grid and the advances of internet of things technology allow for accurate, high resolution measurements of residential PV production as well as high quality, fine-grained weather data such as the forecasts provided by the high-resolution rapid refresh (HRRR) model. 

The performance of models in many sequence-related fields has been recently dramatically improved through the use of modern deep learning (DL) approaches. While this is the case for natural language processing (NLP), language modelling and other similar tasks \cite{s2s, bahdanau, luong}, PV power forecasting has not experienced similar resurrection, as can be seen from current literature reviews~\cite{vandermeer, antonanzas, das, voyant, sobri}. This opens the potential for architectural improvements of DL models in the area of PV power production forecasting.

In this article, we propose and evaluate a sequence to sequence (S2S) model with attention to perform day-ahead forecasts of residential PV power production. This model is based on two seminal innovations from the field of DL for NLP: the introduction of S2S architecture by Sutskever et al. \cite{s2s} and the use of attention mechanisms as additional context generators \cite{bahdanau, luong}. 

We show that this model can leverage high resolution historical data by learning to forecast a time series of binned probability distributions instead of expected values. This increases forecast skill when compared to baseline models and results reported in the current literature.

The remainder of this paper is organized as follows. Section~\ref{sec:bg} presents the technical background concentrating on the PV forecasting practices, sequence related deep learning approaches, and commonly used evaluation metrics. The proposed novel forecasting model is introduced in section~\ref{sec:model} and its performance is evaluated and compared to baseline models in section~\ref{sec:eval}. Final section~\ref{sec:conc} summarizes the results.

\section{Technical Background}\label{sec:bg}

\subsection{PV Power Forecasting Practises}

The PV power forecasting problem has been actively investigated. This research produced a broad variety of forecasting models and tools. They span forecast horizons and resolutions from minutes to weeks, and employ different techniques from physical and statistical modelling to machine learning (ML).
Forecasts are typically provided as expected values of the variable of interest, i.e. a point expected value of PV power production over the forecast step. However, if the forecast were represented as a probability distribution, it would inherently provide more information and thus have a higher value for the user~\cite{vandermeer}. 
Since PV power production depends on multiple variables, it is a common practice to provide the forecasting system with additional data from a numerical weather prediction (NWP) model, including pressure, humidity, temperature, irradiation, and wind speed and direction. And while forecasting the power production from residential installations and from large PV arrays are considered two distinct problems, they largely employ the same techniques. 

A commonly used forecast horizon is 24h ahead with 1h sample step. Since PV power production is highly correlated with solar irradiation, some approaches remove night hours from the data~\cite{sun, math}, or forecast solar irradiation as a proxy~\cite{math}. Additionally, to simplify the PV prediction task, some approaches consider separate predictions on sunny and cloudy days~\cite{days}, or according to seasons~\cite{sun, seasons}. In contrast to such strategies, the proposed model is truly end-to-end: it takes into account all 24h of the forecast and the validation setup contains a randomly selected portion of the data to represent a true measure of performance. 

According to several recent reviews on PV power forecasting \cite{vandermeer, antonanzas, das, voyant, sobri}, the field is dominated by ML approaches. Although algorithms such as support vector regression or random forests are also used, supervised DL is the most frequently applied ML approach. The majority of applied DL architectures include feed-forward neural networks (FFNNs), and recurrent neural networks (RNNs) with different cell types. RNN based models tend to perform better than FFNNs, likely owing to their structure exploiting memory for time series modelling~\cite{lstm}. However, to our best knowledge, none of the reviews~\cite{vandermeer, antonanzas, das, voyant, sobri} nor recent individual models use the concepts of S2S and attention mechanisms that are at the core of the new model proposed in this article. 
        
\subsection{Supervised Deep Learning for Sequences}

Deep learning has been employed, with a great success, in many sequence-related fields -- most notably in  language modelling and NLP. Major advances in performance have been made by abandoning the idea of a one-block model in favor of the encoder-decoder architecture. Before the popularization of this architecture, models were very similar to the one-block designs prevalent in current PV power forecasting literature. As one-block FFNNs do not exhibit a strong, innate autoregressive bias, they are often outperformed by one-block RNNs. 

Of the existing recurrent cells, the long short-term memory cell (LSTM) proposed by Hochreiter and Schmidhuber in~\cite{lstm} is the most common variant for high performing one-block models. The two gates within the LSTM cell allow it to develop contextual memory, making it better suited to extract temporal information patterns from the presented data. Other recurrent cell designs have been proposed and evaluated~\cite{gru}, but so far they failed to challenge the general acceptance of the LSTM. 

The obvious drawback of one-block recurrent architectures is their fixed input-output resolution. For many tasks, such as speech recognition, the input frequency and the output frequency need to be decoupled. And even if a temporal projection layer is used, the performance of one-block recurrent models suffers. 

This led to the proposal of the sequence to sequence (S2S) model by Sutskever et al. \cite{s2s}. It consist of two blocks: one LSTM encoder and one LSTM decoder. The first block encodes the input signal and transfers its final LSTM hidden states to the decoder as a context vector. This allows to collect valuable information from the input sequence, such that the decoder receives only useful features. The decoder then unrolls the output in self-recurrent fashion, feeding its output at time step $t$ back as input at time step $t+1$. This architectural distinction between feature extraction and output synthesis proves to be helpful beyond the decoupling of the input and output time resolutions. 

Bahdanau et al.~\cite{bahdanau} as well as Luong et al.~\cite{luong} both note that S2S has a potential information bottleneck: the decoder cannot access the history of states of the encoder, only the very last pair. To alleviate this, both works propose to use attention mechanism as an additional context generator. 

Attention context of a query $Q$ regarding a key $K$ and a value $V$ can be computed as follows:

\begin{equation}
    \mathrm{score}(Q,K) = \frac{(QW_Q) \times (KW_K)^\intercal}{\sqrt{d_K}}, \label{eq:score}
\end{equation}
\begin{equation}
    A(Q,K,V) = \mathrm{softmax}(\mathrm{score}(Q,K)) \times (VW_V), \label{eq:a}
\end{equation}
where $d_K$ is the length of key $K$, and $W_Q$, $W_K$, $W_V$ are transformations ($W$ denotes a projection into a fixed dimensionality, for example via linearly activated feedforward layer).
$\mathrm{score}(Q,K)$ represents the importance of a feature in $Q$ with respect to $K$. This indicates to which feature the network should pay more attention. The normalization by $\sqrt{d_K}$ is performed to prevent exceedingly large scores that could cause softmax instabilities~\cite{attn}. $A$ is the alignment matrix that represents useful context. 

Both~\cite{bahdanau} and~\cite{luong} show significant improvements over the state of the art in machine translation through the use of attention. These performance gains are mainly attributed to recovered useful context that S2S models otherwise would have difficulties to develop. 

\subsection{Evaluating Forecast Accuracy}
Quality of forecasting models is usually quantified using direct, possibly normalized, error measures. Normalized error measures are preferable, since they provide better comparison between results independent of the size of the PV installation~\cite{vandermeer}. Common direct error measures are (normalized) root mean square error $\mathrm{(n)RMSE}$, (normalized) mean error $\mathrm{(n)ME}$ for point forecasts, and the continuous ranked probability score $\mathrm{CRPS}$ for probabilistic forecasts. For a forecast horizon of $T$ steps $t=1,\ldots,T$ over a target variable with amplitude $P_{\mathrm{max}}$, given a forecast $F$ and real behavior $P$, $\mathrm{nME}$ and $\mathrm{nRMSE}$ are calculated as follows

\begin{equation}
    \mathrm{nME} = \frac{1}{TP_\mathrm{max}} \sum_t |F(t) - P(t)|, \label{eq:nme}
\end{equation}
\begin{equation}
    \mathrm{nRMSE} = \frac{1}{TP_\mathrm{max}} \sqrt{\sum_t \bigl(F(t) - P(t)\bigr)^2}. \label{eq:nrmse}
\end{equation}

If $F(t)$ and/or $P(t)$ are given as probability distributions, the expected values $\expv(F(t))$ and $\expv(P(t))$ are substituted. Within this work, $F(t)$ and $P(t)$ are represented as binned probability distributions over $i_{\mathrm{max}}$ bins. Within the rest of this article, $i$ is omitted in notation unless strictly necessary. To calculate $\mathrm{CRPS}$ for such $F(t,i)$ and $P(t,i)$, their probability density functions (pdf) are first converted to cumulative density functions (cdf) and then compared as follows

\begin{equation}
\mathrm{CRPS} = \frac{1}{i_\mathrm{max}T} \sum_t \sum_i \bigl( F_\mathrm{cdf}(t,i) - P_\mathrm{cdf}(t,i) \bigr)^2. \label{eq:crps}
\end{equation}

The best values of $\mathrm{nRMSE}$ reported in the recent reviews~\cite{vandermeer, antonanzas, das, voyant, sobri} are around 7\%. However, no single set of uniformly adapted metrics exists. As performance is reported on specific data and PV power production exhibits both local and temporal patterns, direct error measurements are not robust. In order to limit local sensitivity, many reviews argue for the use of forecast skill score based on $\mathrm{nRMSE}$ of the model compared to the persistent forecast~\cite{antonanzas} as the preferred performance metric

\begin{equation}
    \mathrm{S}_\mathrm{nRMSE} = 1 - \left( \frac{\mathrm{nRMSE}_\mathrm{model}}{\mathrm{nRMSE}_\mathrm{persistence}} \right). \label{eq:skill}
\end{equation}

The persistent forecast can be defined as the most recent window of the target variable with the same length as the forecast, but without overlap. For a 24h ahead window, this would correspond to the behavior of $P_{-23...0}$. The best models documented in the literature exhibit $\mathrm{S_{nRMSE}}$ of 42.5\% on individual models and 46\% on an ensemble~\cite{pierro}. This article uses $\mathrm{nRMSE}$, $\mathrm{nME}$, $\mathrm{CRPS}$ and $\mathrm{S_{nRMSE}}$, as performance metrics for model evaluation. $\mathrm{S_{nRMSE}}$ is also basis for comparison with other works.

\section{Proposed Model}\label{sec:model}

Instead of one-block models traditionally used for PV power forecasting, we propose to adopt the encoder-decoder approach that previously led to massive improvements of the state of the art in NLP and related fields. Both Bahdanau et al.~\cite{bahdanau} and Luong et al.~\cite{luong} employ attention in encoder-decoder models to maximize the use of available context and improve performance in language related fields. It stands to reason that similar improvement can be expected for forecasting problems.

The encoder-decoder model comes with the benefit of keeping autoregressive bias of recurrent cells while decoupling the input and output frequencies. The high temporal resolution of data provides an opportunity to create more expressive probabilistic targets. The recurrence of the model allows efficient leveraging of these targets for improved accuracy, and the attention mechanism provides additional useful context.

\begin{figure}[t]
    \centering
    \includegraphics[scale=0.5]{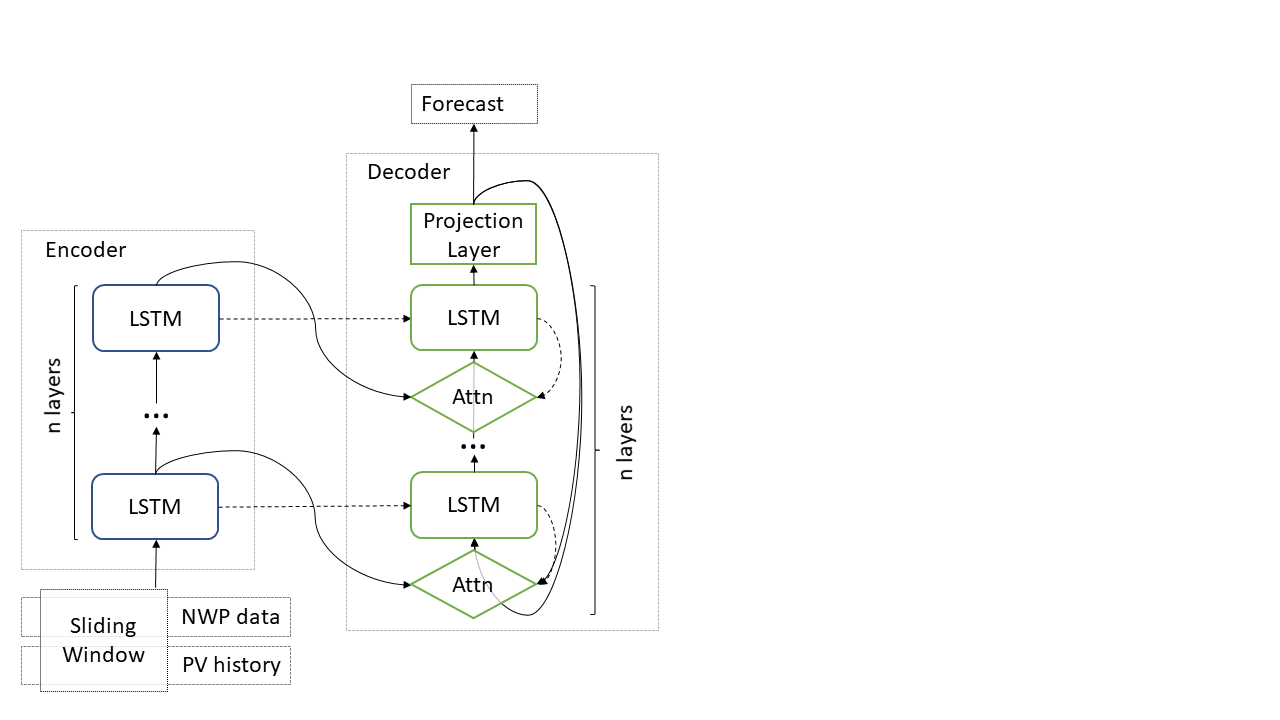}
    \caption{Proposed model \mdl{S2S-Attn}}
    \label{fig:model}
\end{figure}

To test these hypotheses, we develop an LSTM-based encoder-decoder model with attention as additional context generator. The signal flow of the proposed model is depicted in Fig.~\ref{fig:model} and detailed in the following paragraphs.

The encoder consist of $n$ stacked LSTM layers. The input of the encoder is a sliding window $t_{-\mathrm{SW}\ldots0}$ of length $\mathrm{SW}$ up to the current timestep $t_0$. It includes values of NWP forecast and historical PV power consolidated to the same sampling frequency.

After the encoder processes the $\mathrm{SW}$ input, it passes its final states to the decoder as new initial states. The output of the encoder serves as value and key for all attention mechanisms of the decoder. 

The decoder features the same LSTM configuration with the addition of the attention layers. It first performs attention on its input concatenated with the hidden states of the first LSTM layer, then concatenates the resulting context with the input signal, and feeds it into the first LSTM layer. Afterward, there are attention layers preceding the consequent LSTM layers. These layers attend to the hidden states of the corresponding LSTM layer as query and encoder outputs as value, similar to~\cite{luong}. The output of attention is concatenated with the output of the previous LSTM layer and passed to the next. 

The model attempts to learn the underlying distribution $P_\mathrm{true}(t)$ that generates the actual values of PV power, independent of the forecast format, $P(t)$ or $\expv(P(t))$. Therefore, if a sufficient number of values are available to construct an approximation $P(t) \approx P_\mathrm{true}(t)$, it is beneficial to train with $P(t)$ as target because it is closer to the modeled behavior. To avoid bias in constructing such $P(t)$, we chose a binned probability distribution of the output variable over each time interval. Although all models benefit from targets that are closer to the actual behavior in general, the encoder-decoder models benefit further due to their self-recurrence. They perform a rolling forecast, instead of forecasting in one step as in the case of one-block models. One way to leverage this improved information propagation is to learn a more expressive, probabilistic forecast. This allows for more information about past forecast steps to pass on to the next forecast step, freeing up parameters in the model's memory.

To accommodate the probabilistic forecast, the output of the decoder is projected into a binned probability distribution $F(t)$ by a projection layer that features a softmax operator.
In the first decoder step, to forecast $F(1)$, the input is the available $P(0)$. This stands in contrast to similar models for language oriented applications, where the first input is usually a start-of-sentence token. The decoder is trained with teacher forcing, meaning the input received to learn how to forecast $F(t)$ is $P(t-1)$ rather than $F(t-1)$. During evaluation, the decoder is fully self-recurrent and predicts $F(t)$ based on $F(t-1)$. This has been established as the best practice to speed up convergence for encoder-decoder models in language-related fields.

Since the goal of the model is to approximate the probability distribution of $P_\mathrm{true}(t)$ but only an imperfect, observed $P(t)$ is available, the Kullback–Leibler (KL) divergence emerges as a natural choice of loss function. The KL divergence between two binned probability distributions with bins $i$, forecast $F(t,i)$ and true signal $P(t,i)$, can be calculated as

\begin{equation}
    \mathrm{KL}(F,P) = \sum_t \sum_i -P(t,i)\ ln \left( \frac{F(t,i)}{P(t,i)} \right). \label{eq:kl}
\end{equation}

\section{Model Evaluation}~\label{sec:eval}
In summary, previous sections formulate two main hypotheses:
\begin{enumerate}
    \item The proposed model can outperform common one-block models used to forecast PV power and achieve or surpass state of the art performance in terms of the established error measures.
    \item Performance improvements stem from predicting $P(t)$ instead of $\expv(P(t))$ on one hand, and better context extraction through the encoder-decoder architecture with attention and self-recurrence on the other.
\end{enumerate}

\subsection{Experiment Design}
The first claim is evaluated by testing the proposed model, \mdl{S2S-Attn}, against a set of benchmark models. The second claim is quantified by training 2 types of each model: one trained to directly forecast $\expv(P(t))$ over the target time interval, denoted \mdl{model-E}, and one trained to forecast $P(t)$, denoted \mdl{model-pdf}. To further assess the impact of the attention mechanism we also train an attention-less but parameter-equal version of the proposed network, \mdl{S2S}. In summary, the evaluated models are:
\begin{itemize}
    \item \mdl{Persistence}: the persistent forecast model as probabilistic model, using $P(-23...0)$ as $F(1..24)$. This model is used to calculate $\mathrm{S_{nRMSE}}$ and enable comparison with other published models.
    \item \mdl{FFNN-pdf}, \mdl{FFNN-E}: probabilistic and expected value versions of a one block FFNN model.
    \item \mdl{LSTM-pdf}, \mdl{LSTM-E}: probabilistic and expected value versions of a one-block LSTM model.
    \item \mdl{S2S-pdf}, \mdl{S2S-E}: probabilistic and expected value versions of a sequence to sequence model without attention.
    \item \mdl{S2S-Attn-pdf}, \mdl{S2S-Attn-E}: probabilistic and expected value versions of the proposed sequence to sequence model with attention.
\end{itemize}

The PV power data used in the experiments has been provided by  {\em Landmark Homes}. It contains PV power generation values collected with 1 minute resolution at a net zero house in Edmonton between January 2016 and December 2017. The NWP data is a simulation equivalent to the HRRR model, with hourly resolution. It is limited to variables common in PV power forecasting literature: ambient temperature, atmospheric pressure, solar irradiation, wind speed, and relative humidity. The data has been randomly divided into 70\%-15\%-15\% splits for training, testing and validation. The forecast intervals from different sets that overlap are discarded. 

Each model is trained to perform hourly forecasts up to 24 hours ahead. For \mdl{-pdf} models, each hour is a probability distribution with 50 bins from 0 to maximum rated power of the PV installation, while for \mdl{-E} models it is the normalized expected PV power. We attempt to isolate architectural performance influences by keeping the number of parameters, hyperparameter settings and received input data  consistent. Since one-block models cannot process datasteams with different resolution without significant architectural change, we consolidate the two data streams (NWP and PV historical data) to one data stream with  15 min resolution by interpolation and averaging, respectively. For each sample, every model receives a 5-day length equivalent sliding window from this data stream as input. For simplicity the number of units per layer in each model is kept constant. Each \mdl{S2S} model has a 2-layer encoder and a 2-layer decoder. Each \mdl{LSTM} and \mdl{FFNN} one-block model is 2 layers deep and includes a temporal transformation layer to achieve the required output size. More detailed information on benchmark model design and number of parameters can be found in Appendix~\ref{sec:app1}. The training procedure is described in Appendix~\ref{sec:app2}.

As explained in the background section, several error measures ($\mathrm{nRMSE}$, $\mathrm{nME}$, $\mathrm{CRPS}$ and $\mathrm{S_{nRMSE}}$) are employed to provide a detailed overview of model performance. Since some reviewed studies do not employ a separate test set, we report both validation and test performance metrics for clarity. This also allows an estimation of the generalization gap between the development set used to validate and stop training, and the previously unseen set used to perform a true generalization test.

\subsection{Discussion}
\begin{table*}[t]
\caption{Model Performance Comparison}
\begin{center}
\begin{tabular}{|l|c|c|c|c|c|c|c|c|c|c|}
\hline
 &\multicolumn{2}{|c|}{$\mathrm{nRMSE}$}&\multicolumn{2}{|c|}{$\mathrm{nME}$}&\multicolumn{2}{|c|}{$\mathrm{CRPS}$}&\multicolumn{2}{|c|}{$\mathrm{S_{nRMSE}}$}&\multicolumn{2}{|c|}{$S_{\mathrm{CRPS}}$} \\
\cline{2-11} 
\mdl{\textbf{Model}}  &\textit{Val}&\textit{Test}&\textit{Val}&\textit{Test}&\textit{Val}&\textit{Test}&\textit{Val}&\textit{Test}&\textit{Val}&\textit{Test} \\
\hline
\mdl{Persistence} &0.145 &0.133 &0.063 & 0.052 &2.285 &1.944 &- &- &- &- \\
\hline
\mdl{FFNN-E} &0.078 &0.083 &0.054 &0.057 &- &- &0.464 &0.376 &- &- \\
\hline
\mdl{FFNN-{pdf}} &0.072 &0.074 &0.041 &0.040 &1.008 &1.032 &0.501 &0.446 &0.559 &0.469 \\
\hline
\mdl{LSTM-E} &0.073 &0.080 &0.049 &0.054 &- &- &0.496 &0.395 &- &- \\
\hline
\mdl{LSTM-{pdf}} &0.080 &0.087 &0.045 &0.047 &1.166 &1.386 &0.450 &0.344 &0.490 &0.287 \\
\hline
\mdl{S2S-E} &0.089 &0.100 &0.058 &0.065 &- &- &0.388 &0.249 &- &- \\
\hline
\mdl{S2S-{pdf}} &0.068 &0.072 & \textbf{0.039} &0.039 &0.938 &1.003 &0.529 & 0.456 &0.589 &0.484 \\
\hline
\mdl{S2S-Attn-E} &0.119 &0.121 &0.091 &0.094 &- &- &0.184 &0.089 &- &- \\
\hline
\mdl{S2S-Attn-{pdf}} &\textbf{0.067} &\textbf{0.069} &\textbf{0.039} &\textbf{0.038} &\textbf{0.917} &\textbf{0.937} &\textbf{0.536} &\textbf{0.481} &\textbf{0.599} &\textbf{0.518} \\
\hline
\end{tabular}
\label{tab1}
\end{center}
\end{table*}
As can be seen from Table~\ref{tab1}, according to the test set performance, the proposed \mdl{S2S-Attn-pdf} model outperforms all baseline models ($\mathrm{S_{nRMSE}}$ = 48.1\%). The second best performance on the test set is achieved by model \mdl{S2S-pdf} ($\mathrm{S_{nRMSE}}$ = 45.6\%). This confirms the first hypothesis that encoder-decoder models are better suited for a setup with high resolution data. Comparing validation and test performance, \mdl{S2S-Attn-pdf} exhibits a smaller generalization gap then \mdl{S2S-pdf}. This is likely due to the ability of attention to aid generalization through added context.

Another interesting observation is that performance gains from performing probabilistic forecasts are obvious and strong for the tested encoder-decoder models \mdl{S2S-Attn} and \mdl{S2S}. This is not the case for one-block models whose results are less conclusive: model \mdl{FFNN-pdf} outperforms \mdl{FFNN-E}, while \mdl{LSTM-E} outperforms \mdl{LSTM-pdf} for the specific set of hyperparameters listed in the appendix. Thus, it appears that the performance improvement of the encoder-decoder models is not only due to the more expressive, probabilistic target, as the \mdl{-pdf} variants of the one-block models are not consistently affected in the same way. Conversely, the performance improvement cannot solely be attributed to self-recurrence of encoder-decoder models, as \mdl{S2S-Attn-E} and \mdl{S2S-E} significantly under perform. The performance of \mdl{S2S-Attn-pdf} and \mdl{S2S-pdf} therefore stems from the combination of self-recurrence and rich output representation. Latter enables efficient utilization of the former and is especially relevant for $F(1)$, as the probabilistic input $P(0)$ conveys more information about the state of the system at $t=0$ than $\expv(P(0))$. The same argumentation can be applied to the attention mechanism. When compared to model \mdl{S2S-Attn-E}, the probabilistic output of \mdl{S2S-Attn-pdf} provides significantly more features to the first layer's attention mechanism. This allows to construct better attention-based context and results in an additional boost of performance. This validates the second hypothesis from the beginning of this section.

There is no universally agreed upon benchmark dataset for PV forecasting and many authors use custom datasets for experiments. Additionally, our experimental setup necessitates the use of a custom dataset to demonstrate efficient utilization of data with high temporal resolution. Although the proposed model \mdl{S2S-Attn-pdf} is well within the range of state of the art methods concerning the reported direct metrics, comparison based on such metrics is unreliable given the circumstance. Many sources argue that using $\mathrm{nRMSE}$ based Skill compared to a persistent baseline largely alleviates dataset dependency. From reviewed works, the highest reported values of $\mathrm{S_{nRMSE}}$ are between 42.5-46\%~\cite{antonanzas, pierro}. This is significantly lower than the skill of the proposed model \mdl{S2S-Attn-pdf} with $\mathrm{S_{nRMSE}}$ of 53.6\% and 48.1\% for validation and test, respectively. In~terms of $\mathrm{S_{CRPS}}$ the proposed model also features the largest relative increase of this metric (51.8\%) with respect to other models evaluated by van der Meer and Munkhammar~\cite{vandermeer}. However only one work~\cite{bracale} reports a $\mathrm{CRPS}$ improvement over persistence in a scenario similar to that considered in~this study.

Based on all evaluation criteria, we conclude that the proposed model performs at least at and possibly above the state of the art for (residential) PV power forecasting. 

\section{Conclusion}\label{sec:conc}

This article presents motivation for using encoder-decoder models with attention for PV power forecasting. It draws analogy between the state of the art of PV forecasting and the state of the art of NLP tasks before the introduction of these concepts. Examining the flow of information in modern network architectures, we argue that encoder-decoder models are able to leverage high temporal resolution data better than conventional one-block models through two mechanisms: temporal decoupling of encoder and decoder; and decoder self-recurrence. 

To test these hypotheses, we develop and train a sequence to sequence model with attention to perform a probabilistic, binned forecast of PV power production with hourly resolution, up to a day ahead. The input 5-day sliding window data consist of interpolated NWP forecasts and averaged historical PV power data, both with 15 minute resolution. Performance of this model is compared to parameter equivalent sequence to sequence model without attention, to one-block LSTM, and to one-block FFNN. Analogous models that forecast only the expected value are considered as well.

The results suggest that the self-recurrence of the decoder efficiently leverages more expressive, probabilistic targets resulting in a significant performance increase. They also indicate that the temporal decoupling between encoder and decoder leads to a better utilization of the available data. Similarly to models developed in the language processing domain, the proposed model with attention outperforms the attention-less model since it can extract the relevant information from the encoder with a higher efficiency.

Obtained results are compared to other published models based on forecast skill with respect to persistence, an indirect performance measure, well established in the field of PV power forecasting. The proposed sequence to sequence model with attention attains $\mathrm{S_{nRMSE}}$ score of 48.1\% on the test set, outperforming the previously published best skill scores for day-ahead forecasting of 42.5-46\%~\cite{antonanzas, pierro} by a significant margin.

Possible directions for future work include architectural adjustments of the information flow of the encoder-decoder to further specialize the model for forecasting tasks. In addition, it may be possible to adapt other, newer state of the art DL techniques such as self-attention. Finally,  secondary experiments using additional data, possibly from different domains, may further strengthen the validity of results reported in this contribution.

\vspace{12pt}

\clearpage
\appendix
\section{Appendix}
\subsection{Model design}\label{sec:app1}
\begin{figure}[b]
    \centering
    \includegraphics[scale=0.38]{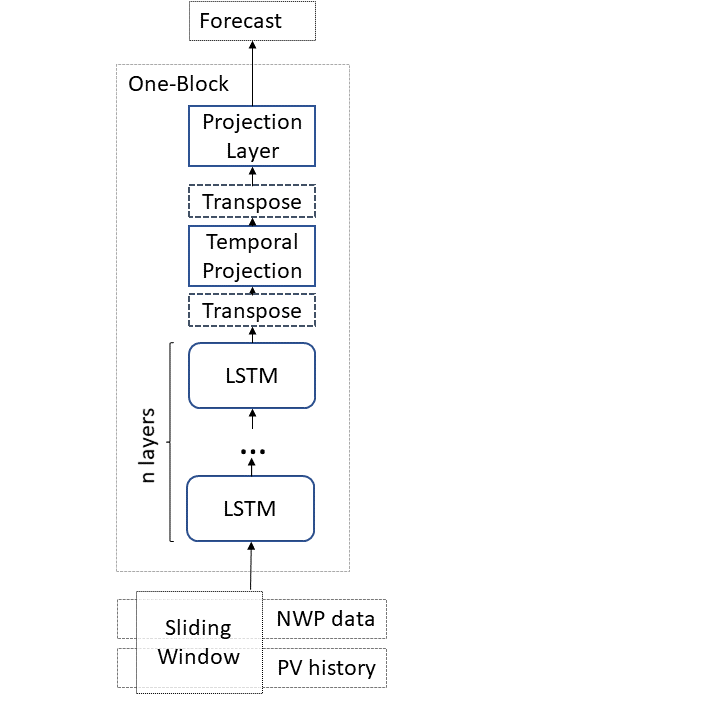}
    \caption{One-block LSTM model}
    \label{fig:1b}
\end{figure}

\begin{figure}[b]
    \centering
    \includegraphics[scale=0.5]{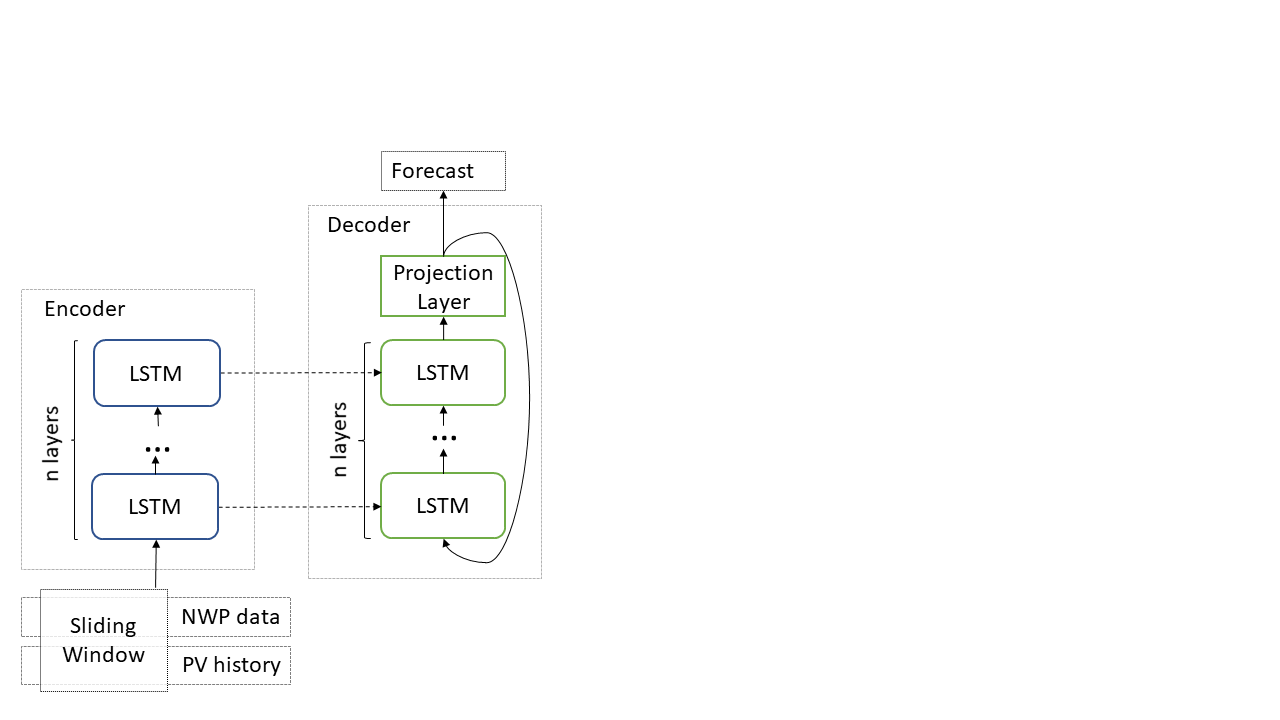}
    \caption{Encoder-decoder model}
    \label{fig:s2s}
\end{figure}

The selected benchmark models include a one-block \mdl{FFNN}, a one-block \mdl{LSTM} (Fig.~\ref{fig:1b}), and a classic encoder-decoder model \mdl{S2S} (Fig.~\ref{fig:s2s}). The \mdl{FFNN} model is constructed similarly to one-block \mdl{LSTM}, but with feedforward layers instead of LSTM. 
One-block models feature a temporal transformation layer. It projects the output of the last layer of the network into the required output shape. In this article, the transformation is a feedforward layer. It reduces the number of timesteps to 24 as per the setup and then transforms the output to the shape of $\expv(P(t))$ or $(P(t))$. 

In order to keep the model size equivalent for all models, the number of units was changed. However, for simplicity, all layers in a given model have the same number of units. Table~\ref{tab2} enumerates the parameters used for model evaluation.

\begin{table}[h]
\caption{Model architecture}
\begin{center}
\begin{tabular}{|l|c|c|}
\hline
\mdl{\textbf{Model}}&\textbf{Number of units per layer}&\textbf{Number of parameters}\\
\hline
\mdl{FFNN-E} &640 &\texttildelow 428k \\
\hline
\mdl{FFNN-{pdf}} &616 &\texttildelow 428k \\
\hline
\mdl{LSTM-E} &184 &\texttildelow 425k  \\
\hline
\mdl{LSTM-{pdf}} &184 &\texttildelow 434k  \\
\hline
\mdl{S2S-E} &132 &\texttildelow 425k \\
\hline
\mdl{S2S-{pdf}} &128 &\texttildelow 431k  \\
\hline
\mdl{S2S-Attn-E} &115 &\texttildelow 441k  \\
\hline
\mdl{S2S-Attn-{pdf}} &110 &\texttildelow 423k \\
\hline
\end{tabular}
\label{tab2}
\end{center}
\end{table}
\subsection{Training procedure}\label{sec:app2}
All models were trained using stochastic gradient descent with a learning rate of 0.003 and nesterov momentum of 0.75. The batch size was 128 and early stopping after 15 epochs of no improvement of $\mathrm{nRMSE}$ was used.

The models forecasting $\expv(P(t))$ were trained using mean squared error ($\mathrm{MSE}$) as a loss function
\begin{equation}
    \mathrm{MSE} = \frac{1}{T} \sum_t (F(t) - \expv(P(t)))^2. \label{eq:mse}
\end{equation}
while $\expv(P(t))$ ranges from 0 to 1.

\end{document}